\documentclass{article}

\usepackage[accepted]{icml2026}

\usepackage[utf8]{inputenc}
\usepackage[T1]{fontenc}
\usepackage{hyperref}
\usepackage{url}
\usepackage{amsfonts}
\usepackage{amsmath}
\usepackage{amssymb}
\usepackage{amsthm}
\usepackage{alltt}
\usepackage{mathtools}
\usepackage{nicefrac}
\usepackage{microtype}
\usepackage{graphicx}
\usepackage{cleveref}
\usepackage{booktabs}
\usepackage{enumitem}
\usepackage{comment}
\usepackage{float}
\usepackage{caption}
\usepackage{balance}
\usepackage[table]{xcolor}

\newcommand{\maci}{MACI}
\newcommand{\ucct}{UCCT}

\usepackage{tikz}
\usetikzlibrary{positioning,arrows.meta}
\begin{document}

% Two-column title page
% REQUIRED: Title must start with "Position:"
\twocolumn[
\icmltitle{AGI Requires a Coordination Layer on Top of Pattern Repositories}

\begin{icmlauthorlist}
\icmlauthor{Edward Y. Chang, Stanford University}{inst1}
\end{icmlauthorlist}

\icmlaffiliation{inst1}{Department of Computer Science, Stanford University}

\icmlcorrespondingauthor{Edward Y. Chang}{echang@cs.stanford.edu}

\icmlkeywords{Large Language Models, Artificial General Intelligence, Multi-Agent Systems, Semantic Anchoring, System-2 Reasoning}
\vskip 0.3in
]

\printAffiliationsAndNotice{}

% REQUIRED: Abstract must identify as position paper
\begin{abstract}
In this paper we argue that influential critiques dismissing Large Language
Models (LLMs) as a dead end for AGI misidentify the bottleneck: they confuse the
ocean with the net. Pattern repositories are the necessary System-1 substrate;
the missing component is a System-2 coordination layer that recruits relevant
patterns, verifies their use, preserves state, and governs convergence. We
separate two uses of control that are often conflated. \emph{Semantic
anchoring}, formalized by \ucct{} (Unified Contextual Control Theory), binds
labels and task intent to learned pattern regions through a phase transition
governed by effective support ($\rho_d$), representational mismatch ($d_r$), and
an adaptive anchoring budget ($\gamma \log k$). \emph{Trace--answer
verification}, implemented by Recursive Causal Audit (RCA), tests whether a
final causal judgment is warranted by its own reasoning trace under pressure. We
translate these ideas into \maci{}, a multi-agent coordination stack that
integrates diversity and control via baiting (PID-modulated debate), filtering
(Socratic and causal audit), and persistence (transactional memory). Empirical
validation on causal judgment and the sycophancy--paranoia trade-off
demonstrates that static prompting fails where adaptive control succeeds. By
reframing common objections as testable coordination failures, we argue that the
path to AGI runs through LLMs, not around them. Capability is not coordination.
\end{abstract}

\section{Introduction: The Field at a Crossroads}
\label{sec:intro}

The artificial intelligence community is fractured by a sharp debate over the
nature of Large Language Models (LLMs) and their role in reaching Artificial
General Intelligence (AGI). Three constructive views dominate. The scaling view
holds that current architectures may be sufficient, needing only more compute,
data, and training refinement~\cite{kaplan2020scaling,hoffmann2022compute,Wei2022EmergentAbilities}.
The reasoning-model view keeps the same substrate but extends it with longer
chains of thought~\cite{Wei2022ChainOfThought} and reinforcement learning on
reasoning traces~\cite{openai2024reasoning,deepseekai2025r1}. The world-model
view argues that the substrate itself must change: AGI requires learned digital
or embodied environments in which agents can simulate, intervene, and collect
new data, as in self-play~\cite{silver2018general}, learned latent
simulators~\cite{ha2018worldmodels}, and JEPA-style architectures~\cite{lecun2022path}.

This third view is paired with an influential critique of the LLM substrate:
LLMs are said to be ``mere pattern matchers,'' structurally incapable of
reasoning, planning, or compositional generalization, and therefore a dead
end~\cite{lecun2025ft,lecun2025newsweek,SutskeverPatel2025ScalingToResearch}.
LeCun states the diagnosis sharply: ``AI systems today are all forms of
System~1,'' and proposes world models as the redesign.

We argue that all three positions identify real ingredients but misplace the
bottleneck. Consider a fishing metaphor. The ocean represents the model's vast
repository of latent patterns. A fisherman casting a net without bait harvests
the \emph{maximum likelihood prior} of the waters beneath them: mostly common
fish (generic training data)~\cite{holtzman2020}. Critics who decry these
ungrounded, hallucinatory outputs are not observing a broken system; they are
observing the raw statistical baseline of an unbaited cast. A world model
enlarges the ocean by adding simulated or embodied dynamics; it does not by
itself decide where to cast, what counts as evidence, or when to stop.

However, intelligent behavior is not just casting; it is \emph{baiting and
filtering}. If the bait is sufficiently dense, it conveys strong intent,
shifting the posterior distribution so that the target concept swamps the common
priors. The ``Missing Layer'' is the coordination layer that optimizes this
trade-off. It transforms the passive retrieval of patterns into the active
construction of reasoning.

\subsection{Our Position: Substrate plus Coordination}

We propose a coordination position: Substrate plus Coordination. We agree that
LLMs alone are insufficient for AGI, but reject the conclusion that they are
irrelevant. 

\textbf{Our central thesis is}:

\vspace{0.5em}
\textit{LLMs are the necessary System-1 substrate: the pattern repository. The
primary bottleneck for AGI is not the absence of patterns, but the absence of a
System-2 coordination layer that binds patterns to constraints, verifies their
use, preserves state, and regulates convergence.}
\vspace{0.5em}

The System-1/System-2 distinction must be stated precisely~\cite{kahneman2011},
because our position rests on it. System-1 is fast, reflexive, automatic
processing: pattern completion without explicit deliberative control. System-2 is slower, deliberate processing that proceeds through explicit reasons,
verification, memory, and control. An LLM is System-1: a pattern repository
formed by next-token prediction under maximum likelihood. We accept this
diagnosis: LeCun's claim that today's models are ``all
System-1''~\cite{lecun2025ft,lecun2025newsweek} is correct. Crucially, the
layers trained on top of this substrate (instruction tuning, chain-of-thought, and reinforcement learning on reasoning traces) do not change the stratum. They reshape the pattern distribution and produce fluent simulations of deliberation,
but they do not supply the coordination layer needed for governed reasoning.
This is why, post-fine-tuning, models still exhibit
sycophancy~\cite{sharma2023sycophancy}, unfaithful reasoning
traces~\cite{turpin2023unfaithful,lanham2023measuring}, and the failure modes
cataloged in \S\ref{sec:empirical}.

Our position follows directly. The LLM substrate is \emph{necessary but
insufficient}: necessary because System-2 has nothing to deliberate over without
a rich repository of patterns; insufficient because pattern association alone
does not become governed reasoning. The missing component is System-2 itself.
And System-2 is not a single mechanism 
bolted onto the substrate; like human deliberative cognition, it is a layered stack with structure of its own.

This stack comprises four control levels, each more global than the last.
\emph{Level~1: Semantic Anchoring} recruits relevant patterns, binding labels
and intent while suppressing high-prior drift. \emph{Level~2: Trace--Answer
Verification} tests whether a trace warrants its answer. \emph{Level~3:
Persistence} records commitments, reasons, contradictions, and rollbacks across
episodes. \emph{Level~4: Governance} coordinates agents, regulates convergence,
and decides when to assert, revise, or refuse. These are not disjoint systems:
\maci{} is the integrative architecture composing all four, with \ucct{}, RCA,
transactional memory (SagaLLM~\cite{sagallm2025}), and \maci{} governance
instantiating Levels~1--4. The path to AGI is to build this stack above the
substrate, not to choose between scaling and abandonment.

Figure~\ref{fig:coordination_hierarchy} summarizes this organization: a System-1
substrate layer supports a System-2 coordination layer composed of four
increasingly global control levels: semantic anchoring, trace--answer
verification, persistence, and governance.

\begin{figure}[th]
\centering
\resizebox{\columnwidth}{!}{%
\begin{tikzpicture}[
    font=\scriptsize,
    rowbox/.style={
        draw=black!70,
        rounded corners=2pt,
        minimum width=7.45cm,
        minimum height=0.90cm,
        inner sep=3pt
    },
    headerbox/.style={
        draw=blue!55!black,
        fill=blue!10,
        rounded corners=2pt,
        minimum width=7.45cm,
        minimum height=0.38cm,
        inner sep=2pt,
        font=\scriptsize\bfseries,
        align=center
    },
    subheader/.style={
        draw=gray!70,
        fill=gray!18,
        rounded corners=2pt,
        minimum width=7.45cm,
        minimum height=0.38cm,
        inner sep=2pt,
        font=\scriptsize\bfseries,
        align=center
    },
    lefttext/.style={
        font=\scriptsize\bfseries,
        align=center,
        text width=1.95cm
    },
    righttext/.style={
        font=\scriptsize,
        align=left,
        text width=5.00cm
    }
]
\newcommand{\coordrow}[4]{%
    \node[rowbox, fill=blue!4] (#1) at (0,#2) {};
    \draw[black!45] (-1.65,#2-0.38) -- (-1.65,#2+0.38);
    \node[lefttext] at (-2.70,#2) {#3};
    \node[righttext] at (1.04,#2) {#4};
}
% System-2 coordination layer
\node[headerbox] at (0,4.50) {System-2 Coordination Layer};
\coordrow{l4}{3.85}{Level 4\\Governance}
{multi-agent control; convergence; refusal\\
\emph{MACI, RCA}}
\coordrow{l3}{2.90}{Level 3\\Persistence}
{transactional memory; commitments; rollback\\
\emph{MACI memory}}
\coordrow{l2}{1.95}{Level 2\\Trace--Answer\\Verification}
{trace--answer audit; critique; causal checking\\
\emph{RCA, Socratic judging}}
\coordrow{l1}{1.00}{Level 1\\Semantic\\Anchoring}
{label/intent binding; drift suppression\\
\emph{UCCT}}
% System-1 substrate layer
\node[subheader] at (0,0.10) {System-1 Substrate Layer};
\node[rowbox, fill=gray!10] (substrate) at (0,-0.65) {};
\node[font=\scriptsize\bfseries, align=center, text width=1.95cm] 
    at (-2.70,-0.65) {Substrate};
\node[font=\scriptsize, align=left, text width=5.00cm] 
    at (1.04,-0.65) {pretrained LLM pattern repository\\
    \emph{associations among learned patterns}};
\end{tikzpicture}}
\caption{The proposed hierarchy of control. The pretrained LLM supplies the System-1 substrate; the System-2 coordination layer organizes semantic anchoring, trace--answer verification, persistence, and governance above that substrate.}
\label{fig:coordination_hierarchy}
\vspace{-.1in}
\end{figure}

The contrast with reasoning models is instructive. Such models fine-tune the
LLMs with ever-longer chains of thought, yet the empirical record shows the
pathologies sharpening rather than receding. As reasoning training intensifies
across the GPT-5 line, sycophancy \emph{rises} with capability rather than
falling, reaching 11.4\% for GPT-5.1, and counterfactual (Pearl L3) Safety
\emph{collapses} to 20\% for the most capable model (\S\ref{sec:empirical}). A
more capable LLM simulates reasoning more fluently, which includes
constructing more fluent rationalizations for wrong answers. In the fishing
metaphor, a reasoning model is a larger, better-stocked ocean and a longer cast;
it is still neither the net nor the bait. Capability is not coordination.

This paper formalizes and instantiates this hierarchy. At the theoretical level,
we present \ucct{} (Unified Contextual Control Theory), which models reasoning as
a phase transition governed by anchoring strength~\cite{ucct2025}. At the
architectural level, we introduce \maci{} (Multi-Agent Collaborative
Intelligence), a disciplined control system for multi-agent reasoning. Unlike the
current wave of agentic hype, which often produces chaotic or divergent
swarms, \maci{} offers a \textbf{totality view}: diversity of perspective is
valuable only when regulated toward \textit{reasoned convergence}. Its core
mechanisms are regulated debate (baiting), Socratic judging (filtering),
transactional memory (state), and governance policies for convergence and
refusal~\cite{path2AGI-V1}.

At the empirical level, we use Recursive Causal Audit (RCA) to show that
reliability requires adaptive control~\cite{ChangACL2026}. The distinction is
precise: \ucct{} provides semantic anchoring, binding meanings and intent to
learned pattern regions; RCA provides trace--answer verification, checking
whether a final causal judgment is warranted by the model's derivation,
internally consistent, and resistant to pressure-induced hint adoption. Without
PID-style regulation, models oscillate between \emph{Sycophancy}: under-control,
where they capitulate to users, and \emph{Paranoia}, over-control, where they
reject valid reasoning. This control landscape supports the central claim:
failures attributed to substrate limitations are better understood as failures
of coordination.

\paragraph{A compact operational lens.}
We summarize anchoring with a scalar strength $S=\rho_d-d_r-\gamma\log k$:
effective support minus representational mismatch minus an adaptive
context-budget penalty. Section~\ref{sec:phase} derives it in full.

This paper unifies two strands of our broader AGI program. Volume~I developed
\maci{} as a theory of multi-LLM collaborative intelligence: roles, debate,
critique, and transactional memory~\cite{path2AGI-V1}. Volume~II develops the
System-2 account: semantic anchoring, causal validation, and recursive
audit~\cite{path2AGI-V2}. Together, they define the coordination
hierarchy that turns a pretrained pattern repository into a governed reasoner.

\subsection{Why This Matters}
This distinction determines what the field optimizes. If the substrate is
broken, as the ``Dead End'' view suggests, then the field must discard LLMs and
invent new foundations. If the substrate is necessary but incomplete (lacking
the System-2 stack), as we argue, then the engineering priority changes: build
the four control levels (anchoring, verification, persistence, governance) that
transform pretrained capacity into reliable inference.

\paragraph{Roadmap.}
Section~\ref{sec:biological} grounds the coordination view in biological
cognition; Section~\ref{sec:phase} derives the \ucct{} phase transition;
Section~\ref{sec:maci} details the \maci{} architecture; and
Section~\ref{sec:empirical} presents RCA evidence for control-theoretic
coordination. An extended positioning against the debate, verification,
agentic, and control-theory literatures is given in
Appendix~\ref{app:extended-related}.

\section{The Biological Foundation: Intelligence Emerges From Pattern Repositories}
\label{sec:biological}

To evaluate claims that LLMs are ``mere pattern matching,'' we start from biological cognition. Intelligence does not oppose pattern-based processing; higher-level deliberation is built by organizing and regulating large pattern repositories.

%\vspace{-.08in}
\subsection{Unconscious Cognition: The ``Ocean'' of Substrate}

A substantial fraction of human competence is implemented by fast, specialized subsystems operating below awareness. In our fishing metaphor, these systems constitute the ocean: a vast population of latent behaviors and priors.

These include autonomic regulation, threat appraisal, motor control, hierarchical perception, and automatic language processing~\cite{kandel2013,gazzaniga2014}. These systems are not peripheral; they are the substrate that makes everyday reasoning possible. It is plastic: practice continually adds structure, expanding what can be executed quickly and reliably~\cite{shiffrin1977}.

In LLMs, mechanistic interpretability reveals analogous structure: circuits implementing specific computations~\cite{olah2020zoom} and superposition enabling high-dimensional feature encoding~\cite{elhage2022toy}. Without coordination, sampling from this repository yields degenerate text: repetitive, generic, or incoherent outputs reflecting the maximum likelihood prior~\cite{holtzman2020}.

\begin{figure}[t]
    \centering
\includegraphics[width=1.0\linewidth, height=0.19\textheight]{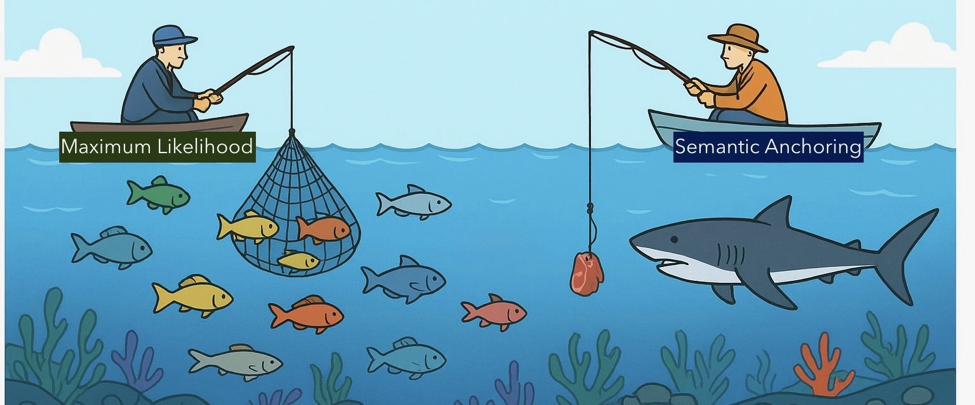}
    \caption{\textbf{The Mechanics of Coordination.} \textbf{Left:} Without anchors, the model retrieves high-prior tokens. \textbf{Right:} Semantic bait shifts support ($\rho_d$), capturing rare goal-directed targets otherwise drowned by priors.}
    \label{fig:fishing_metaphor}
    \vspace{-.15in}
\end{figure}

%\vspace{-.08in}
\subsection{Conscious Control: The ``Net'' and the ``Bait''}

Conscious, goal-directed reasoning (System-2) is slower and more resource-limited~\cite{kahneman2011}. A common mistake is to treat this as evidence for a fundamentally different computational basis. Bengio~\cite{bengio2019consciousness} proposes the ``consciousness prior'', the idea that conscious processing selects and combines unconscious representations into coherent, verbalizable thoughts. A better view is that conscious control operates by \emph{fishing} from the underlying repositories: selecting, constraining, and organizing specific patterns.

The prefrontal cortex implements this coordination, and its functions map onto
the four-level stack of Section~\ref{sec:intro}. Deliberate problem solving
queries stored patterns by broadcasting goals, the \emph{bait} that recruits a
target region (Level~1, semantic anchoring). Executive function then imposes
constraints, the \emph{mesh} that filters the catch, rejecting unstable or
unsupported candidates (Level~2, trace--answer
verification)~\cite{gazzaniga2014}. Working memory holds intermediate
commitments and prior conclusions across a deliberation, so that reasoning can
be revised rather than restarted (Level~3,
persistence)~\cite{baddeley1992working}. Metacognitive monitoring decides when
a conclusion is warranted, when to revise it, and when to withhold judgment
under uncertainty (Level~4, governance)~\cite{flavell1979metacognition}.
System-2 is not a non-pattern-based alternative; it is this layered coordination
regulating which patterns surface, which persist, and which govern action.

\subsection{Sharp Transitions: Empirical Evidence}

Our \ucct{} experiments show that semantic anchoring produces sharp, abrupt
behavior. Two cases make the point. \emph{Subtraction override:} two in-context
examples that redefine ``$-$'' as ``$+$'' (7\,$-$\,4\,$=$\,11;
5\,$-$\,2\,$=$\,7) flip every tested frontier LLM from 5 to 11 on ``$8-3$.''
\emph{Underdetermined patterns:} when examples admit several consistent rules,
models split (23 vs.\ 70 on ``$15-8$''), revealing that the prior, not robust
reasoning, selects the rule. Across demonstrations, behavior tracks the Max-$S$
comparison of Section~\ref{sec:phase}: the prior wins with too few anchors, the
anchored target wins once its score overtakes the prior, and over-baiting
reverses the gain. Full details are in Appendix~\ref{app:demonstrations}.

% ============================================================
% Begin inlined file: PhaseTransition.tex
% ============================================================

%%%%%%%%%%%%%%%%%%%%%%%%%%%%%%%%%%%%%%%%%%%%%%%%%%%%%%%%%%%%%%%%%%%%%%%%
%\vspace{-0.1in}
\section{Phase Transitions: Patterns to Cognition}
\label{sec:phase}
%%%%%%%%%%%%%%%%%%%%%%%%%%%%%%%%%%%%%%%%%%%%%%%%%%%%%%%%%%%%%%%%%%%%%%%%

\begin{figure}[t]
    \centering  \includegraphics[width=1.0\linewidth, height=0.25\textheight]{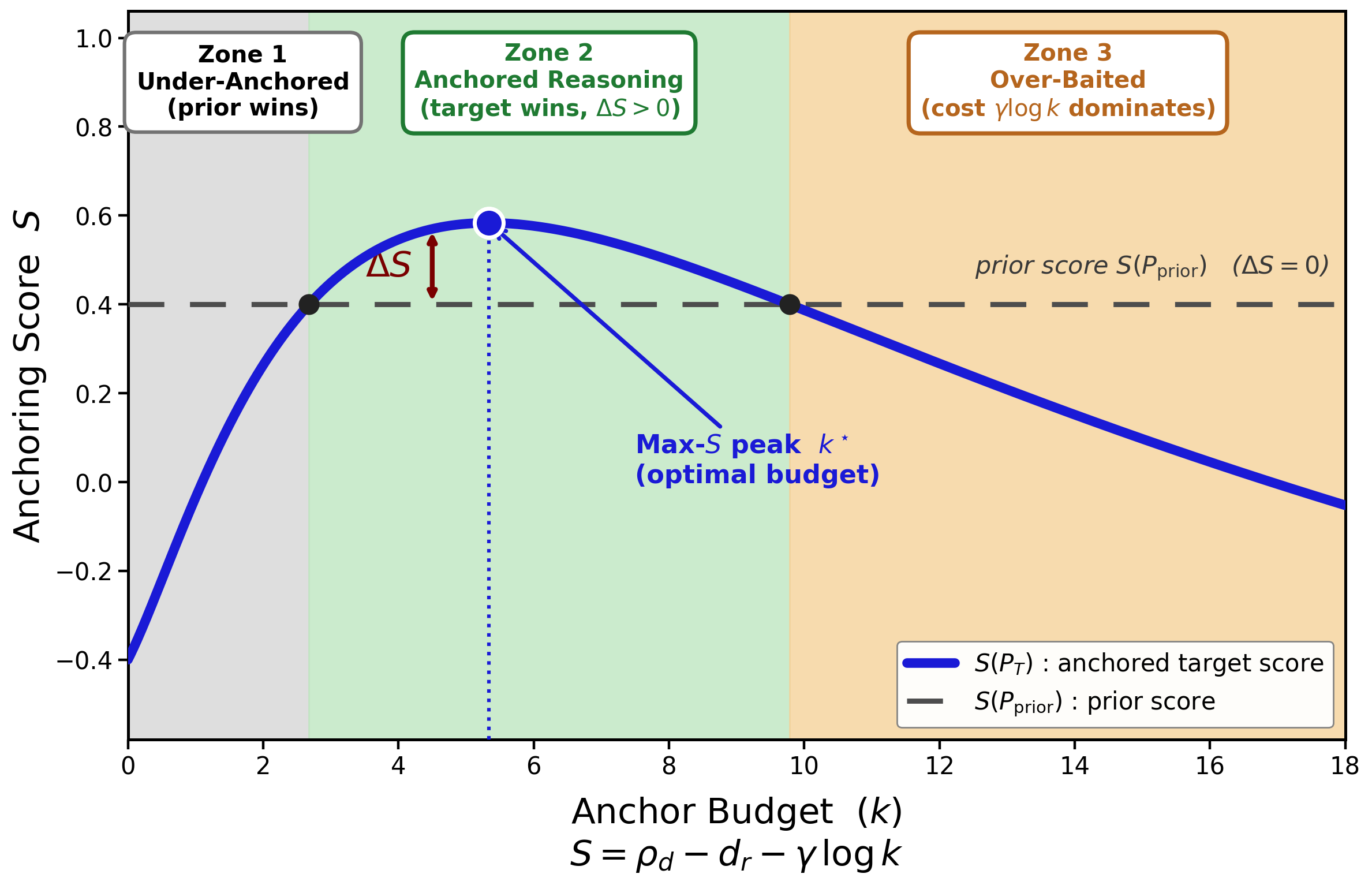}
\caption{\textbf{The Physics of Coordination.} Anchored behavior follows
Max-$S$ selection: the model runs the target rule $P_T$ only while its score
$S(P_T)$ (blue) exceeds the prior's $S(P_{\mathrm{prior}})$ (dashed), that is,
while $\Delta S>0$. Adding anchors raises $S(P_T)$ through effective support
$\rho_d$, but the budget cost $\gamma\log k$ grows without bound, so $S(P_T)$
peaks at an optimal budget $k^\star$ and then declines. \textbf{Zone~1:} too
little bait, the prior wins. \textbf{Zone~2:} anchored reasoning, $\Delta S>0$.
\textbf{Zone~3:} over-baiting, the budget cost reclaims the prior. No external
threshold is posited.}
\label{fig:phase_transition}
\vspace{-0.1in}
\end{figure}

Section~\ref{sec:biological} showed that a tiny amount of context can override
an enormous pretrained repository, producing abrupt behavioral flips. This is
not a machine-learning oddity but a universal mechanism: \emph{abrupt state
change}. Figure~\ref{fig:phase_transition} illustrates the core idea: as the
anchor budget grows, the target rule's score $S(P_T)$ rises, overtakes the
pretrained prior, peaks at an optimal budget, and can fall again when
over-baiting makes context too costly. Large pattern repositories are not a
dead end; they are the substrate that makes this anchor-driven reconfiguration
possible.

\subsection{Abrupt transitions are ubiquitous}

Abrupt transitions arise when a system has multiple stable regimes separated by an effective barrier, and when feedback amplifies deviations near a critical point. In physics, canonical examples include liquid--gas transitions at a boiling point (for fixed pressure), ferromagnetic ordering at the Curie temperature, and the onset of superconductivity below a critical temperature. In biology, the same qualitative structure appears in action potentials: membrane voltage integrates smoothly until a threshold triggers an all-or-nothing spike. Switch-like behavior also arises in gene-regulatory networks, where positive feedback yields commitment to a developmental fate. 

In machine learning, Power et al.~\cite{power2022grokking} discovered
``grokking,'' delayed generalization that emerges suddenly after extended
training on algorithmic tasks. Nanda et al.~\cite{nanda2023progress} provide
mechanistic interpretability of this phenomenon, showing that internal
representations undergo qualitative reorganization at the transition point.
These examples differ in medium and scale, yet share a common signature: near a
critical point, small quantitative changes can trigger a qualitative change in
state.

\subsection{\ucct{}: semantic anchoring as a cognitive phase transition}

\ucct{} makes the phase-transition interpretation explicit for LLM behavior under external structure. It posits a scalar \emph{anchoring strength} that summarizes when external structure successfully binds to latent patterns.

\paragraph{Anchoring strength with adaptive regularization.}
The score plotted in Figure~\ref{fig:phase_transition} is anchoring strength, defined as:
\begin{equation}
S \;=\; \rho_d \;-\; d_r \;-\; \gamma \log k,~\quad \text{where:}
\label{eq:anchoring_score}
\end{equation}

\begin{itemize}[itemsep=1pt,topsep=1pt,leftmargin=1.2em]
    \item \textit{Effective Support ($\rho_d$):} The \emph{density of the bait}. This measures how strongly the current cues recruit the target concept in the latent space. If $\rho_d$ is sparse (weak cues), the signal fails to overcome the model's training priors.
    \item \textit{Mismatch ($d_r$):} The mesh size of the net. This captures the instability of the representation under perturbation; a finer mesh (low $d_r$) filters out hallucinations and unstable candidates.
    \item \textit{Adaptive Regularizer ($\gamma \log k$):} The cost of the bait. While increasing the amount of bait ($k$) generally increases density ($\rho_d$), it incurs a cost. If $\gamma$ is high (e.g., in a noisy or resource-constrained environment), the system penalizes ``over-baiting,'' enforcing the cognitive reality that efficient intelligence must solve problems without unbounded context.
\end{itemize}

\paragraph{Measurement recipe.}
Each term of Eq.~\eqref{eq:anchoring_score} is estimated from observable
quantities: the anchoring budget $k$ counts admitted anchors, mismatch $d_r$
comes from output sensitivity under perturbation, and effective support
$\rho_d$ from self-consistency across sampled reasoning paths. This makes 
$S$ empirically testable; full measurement formulas are given in the companion UCCT paper~\cite{ucct2025}.

\paragraph{Max-$S$ selection: the threshold is derived, not posited.}
\ucct{} does not posit a critical value that $S$ must exceed. Behavior is
governed by \emph{Max-$S$ selection}. For a given query, the candidate latent
rules are the target rule $P_T$ that the anchors recruit and the pretrained
prior $P_{\mathrm{prior}}$; the model executes whichever carries the larger
score. Behavior tips when the target overtakes the prior, that is, when the gap
\begin{equation}
\Delta S \;=\; S(P_T) \;-\; S(P_{\mathrm{prior}})
\label{eq:phase}
\end{equation}
changes sign. The decision boundary is  $\Delta S = 0$: a
comparison between two scores, not a fitted parameter. The budget term
$-\gamma\log k$ is part of the score being maximized, a BIC-style penalty on
anchor spend, not a separate cutoff.

\paragraph{Over-baiting: more anchors are not always better.}
The budget cost $\gamma\log k$ grows without bound while effective support
$\rho_d$ saturates, so $S(P_T)$ is not monotone in the anchor budget.
Figure~\ref{fig:phase_transition} traces this: $S(P_T)$ rises as anchors
accumulate, peaks at an optimal budget $k^\star$, then declines, because past
$k^\star$ additional context costs more than it contributes. Three regimes
follow from the Max-$S$ comparison. With too few anchors the target never
overtakes the prior (Zone~1). In an intermediate band the target wins,
$\Delta S>0$ (Zone~2). With excessive anchoring the cost term pulls $S(P_T)$
back below the prior, and the prior reclaims control (Zone~3). The transition
is set by a comparison of two scores, with no external threshold.

\paragraph{Why pattern repositories are necessary.}
A rich substrate is not a defect but the enabling condition for anchor-driven
regime change. Dismissing LLMs as ``mere pattern matching'' misses what pattern
repositories make possible: discrete regime shifts under small, structured
interventions.

% ============================================================
% End inlined file: PhaseTransition.tex
% ============================================================

% ============================================================
% Begin inlined file: CatQuery.tex
% ============================================================

\section{Example: The Four-Year-Old's Cat}
\label{sec:cat}

A four-year-old learns to identify cats from 3-4 labeled photos. This is not rote memorization but a sharp transition: a small amount of labeled context recruits a large pre-existing repository, producing a qualitative shift from diffuse similarity to stable category identification. The same mechanism operates in LLMs.

\begin{figure}[t!]
\centering
\begin{tikzpicture}
% === TOP ROW: High rho_d (green borders) ===
\node[inner sep=0pt] (cat) at (0.7, 1.5) 
    {\includegraphics[width=2.0cm, height=1.4cm]{Figs/Easy-cats.jpeg}};
\draw[green!60!black, line width=1.5pt] (cat.south west) rectangle (cat.north east);
\node[below=2pt of cat, font=\footnotesize] {Cat};

\node[inner sep=0pt] (dog) at (2.8, 1.5) 
    {\includegraphics[width=2.0cm, height=1.4cm]{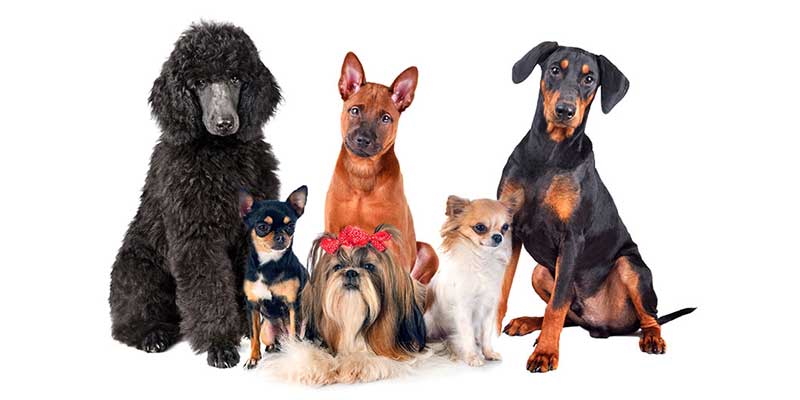}};
\draw[green!60!black, line width=1.5pt] (dog.south west) rectangle (dog.north east);
\node[below=2pt of dog, font=\footnotesize] {Dog};

\node[inner sep=0pt] (elephant) at (4.9, 1.5) 
    {\includegraphics[width=2.0cm, height=1.4cm]{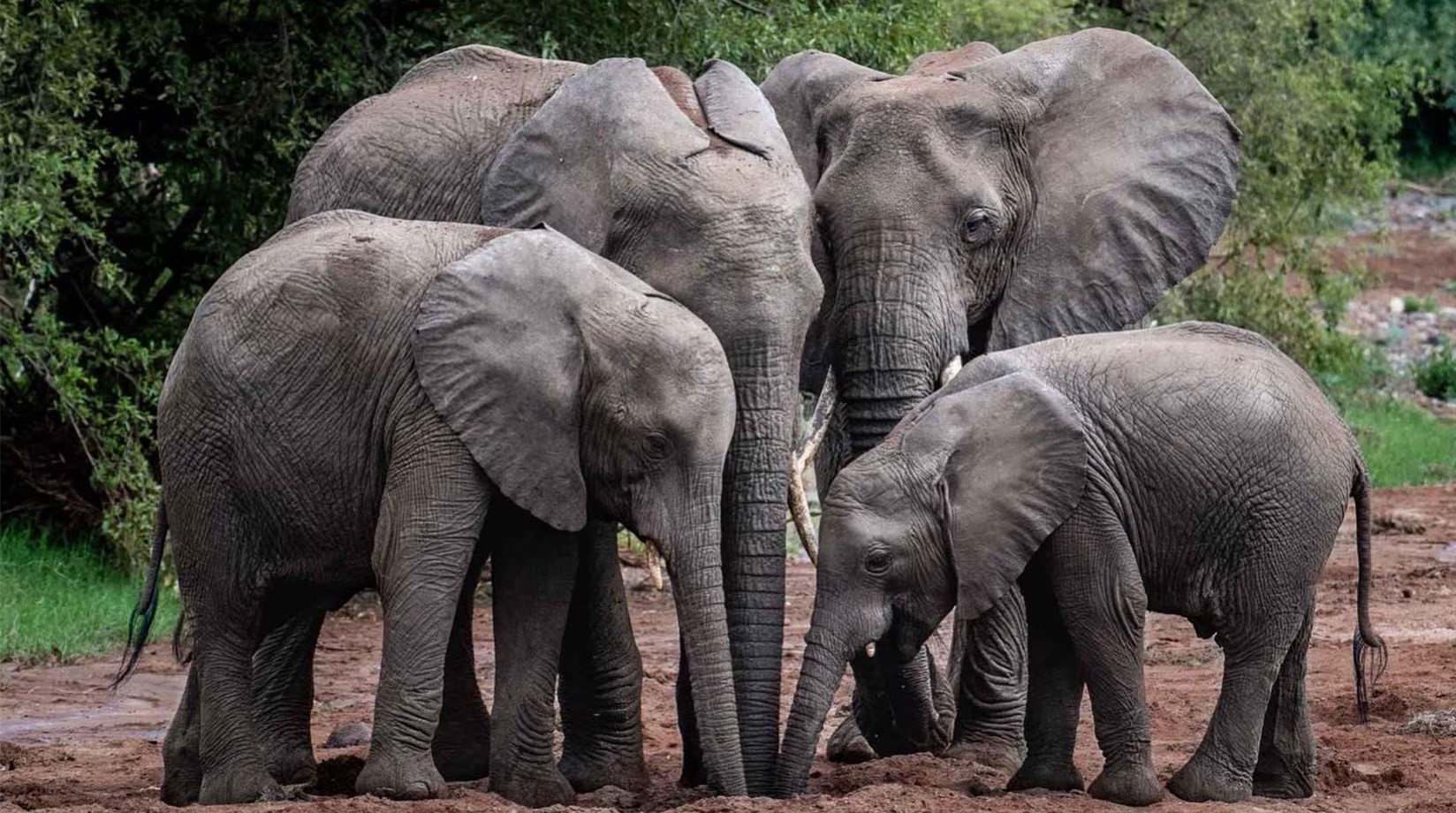}};
\draw[green!60!black, line width=1.5pt] (elephant.south west) rectangle (elephant.north east);
\node[below=2pt of elephant, font=\footnotesize] {Elephant};

\node[inner sep=0pt] (bird) at (7.0, 1.5) 
    {\includegraphics[width=2.0cm, height=1.4cm]{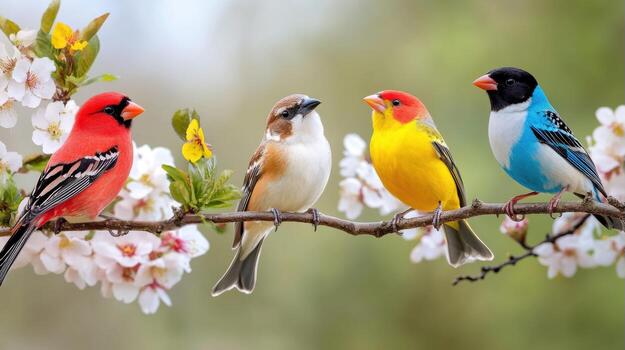}};
\draw[green!60!black, line width=1.5pt] (bird.south west) rectangle (bird.north east);
\node[below=2pt of bird, font=\footnotesize] {Bird};

% === BOTTOM ROW: Low rho_d (red borders) ===
\node[inner sep=0pt] (pangolin) at (0.7, -0.5) 
    {\includegraphics[width=2.0cm, height=1.4cm]{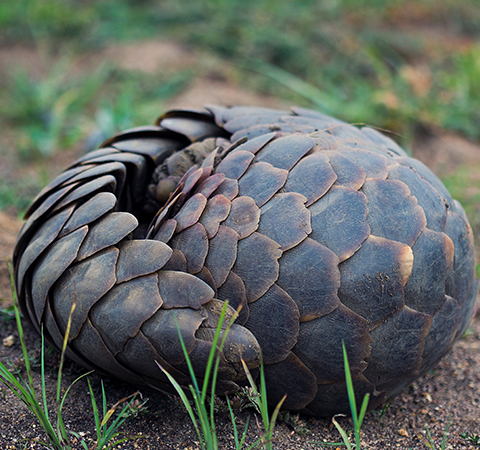}};
\draw[red!70!black, line width=1.5pt] (pangolin.south west) rectangle (pangolin.north east);
\node[below=2pt of pangolin, font=\footnotesize] {Pangolin};

\node[inner sep=0pt] (okapi) at (2.8, -0.5) 
    {\includegraphics[width=2.0cm, height=1.4cm]{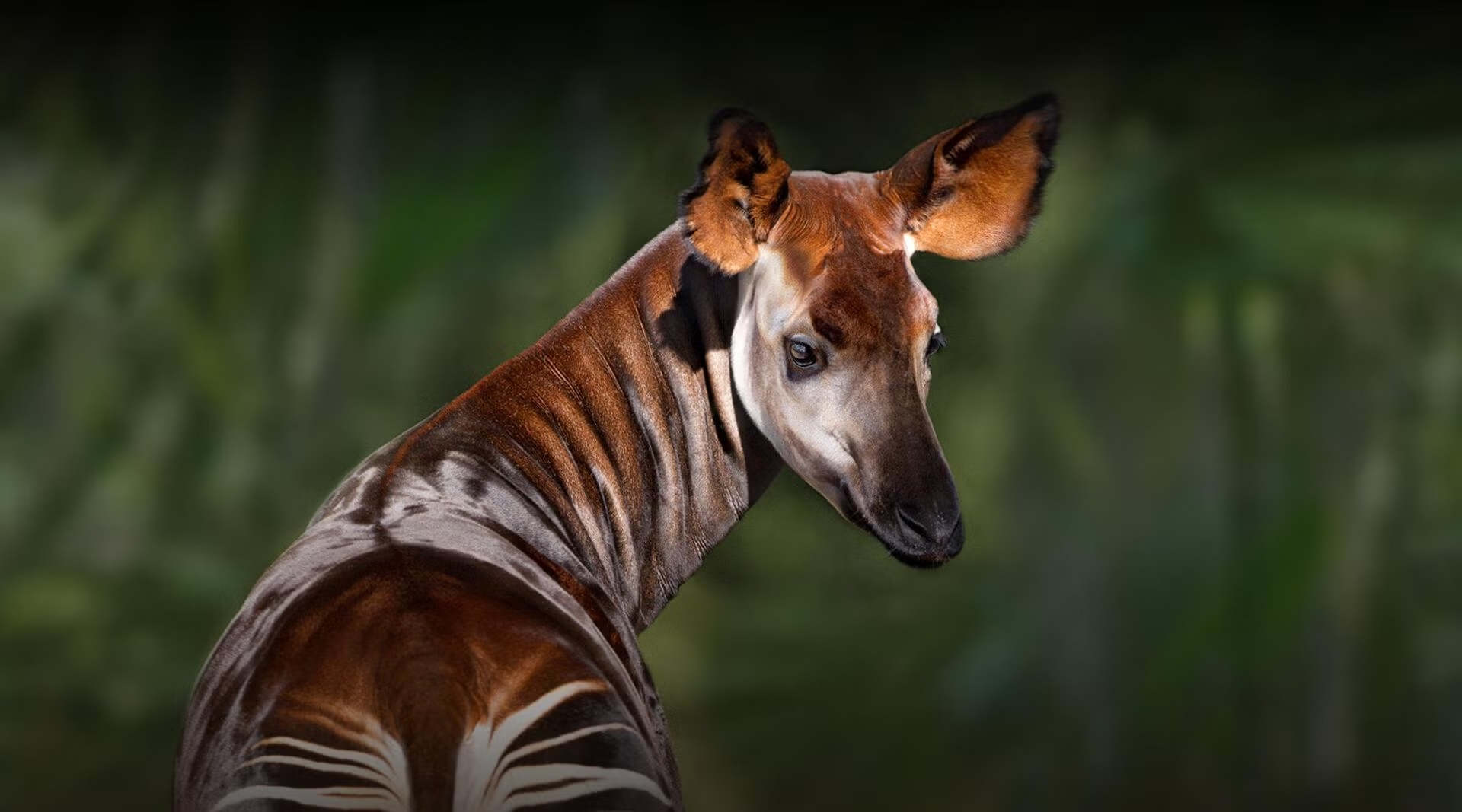}};
\draw[red!70!black, line width=1.5pt] (okapi.south west) rectangle (okapi.north east);
\node[below=2pt of okapi, font=\footnotesize] {Okapi};

\node[inner sep=0pt] (axolotl) at (4.9, -0.5) 
    {\includegraphics[width=2.0cm, height=1.4cm]{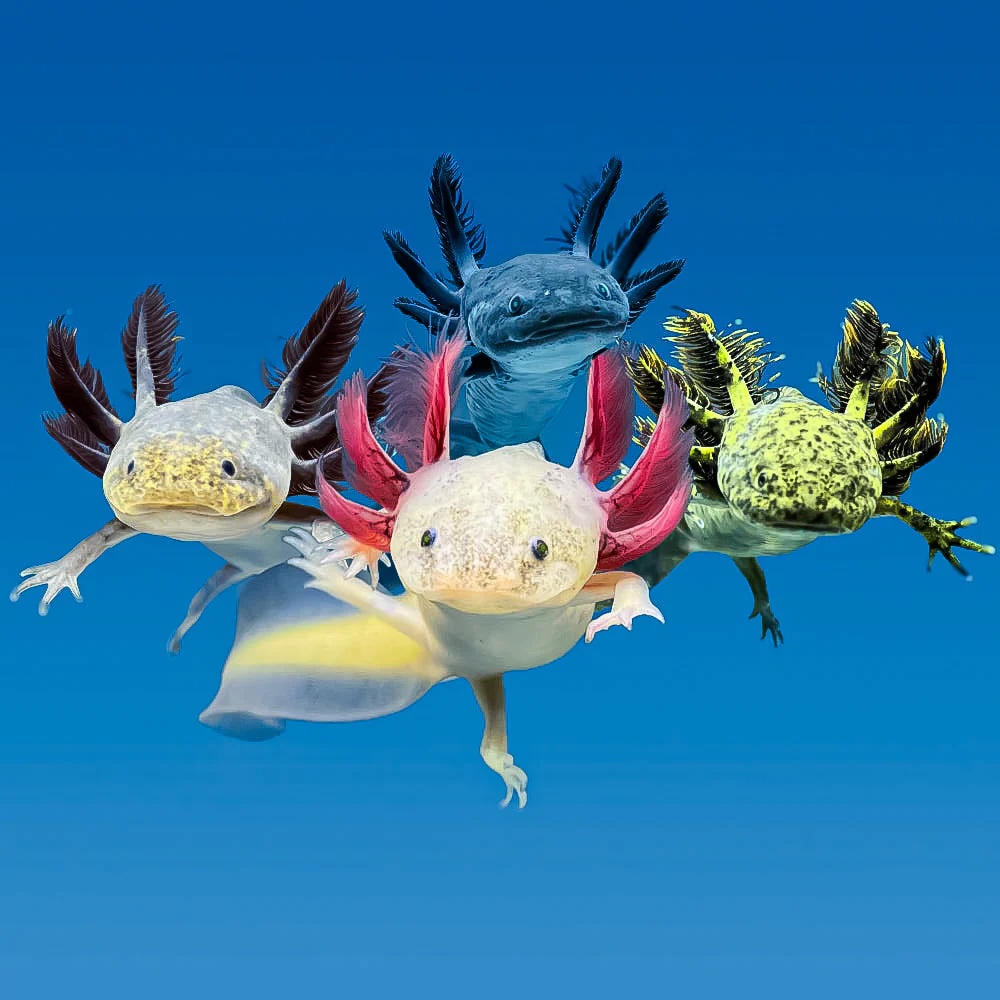}};
\draw[red!70!black, line width=1.5pt] (axolotl.south west) rectangle (axolotl.north east);
\node[below=2pt of axolotl, font=\footnotesize] {Axolotl};

\node[inner sep=0pt] (tarsier) at (7.0, -0.5) 
    {\includegraphics[width=2.0cm, height=1.4cm]{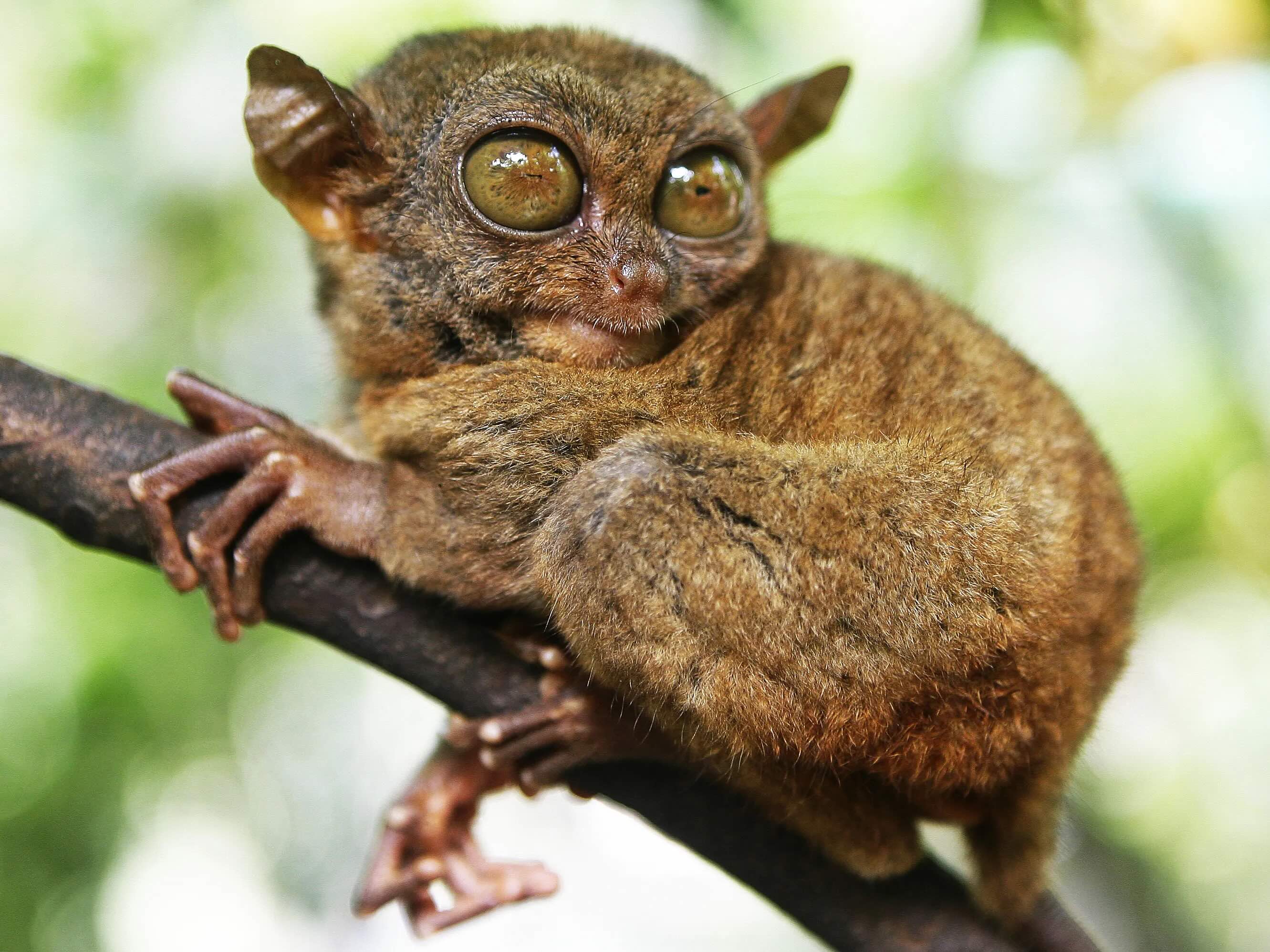}};
\draw[red!70!black, line width=1.5pt] (tarsier.south west) rectangle (tarsier.north east);
\node[below=2pt of tarsier, font=\footnotesize] {Tarsier};
%\vspace{-.1in}
\end{tikzpicture}
\caption{Pattern density determines anchoring difficulty. Familiar animals (top) anchor in 3--4 shots; rarely seen animals (bottom) may fail even with 10 examples. Difficulty depends on specificity: a child identifies ``dog'' easily but cannot name breeds; ``pangolin'' fails at any level without prior exposure; ``okapi'' gets mapped to ``zebra'' or ``deer.''}
\label{fig:animal_contrast}
\vspace{-.1in}
\end{figure}

%\vspace{-0.08in}
\paragraph{The parallel.}
Table~\ref{tab:child_llm} makes the analogy explicit. Both systems accumulate massive pattern repositories \emph{before} labeled examples arrive: the child through 4 years of looking around and watching TV, the LLM through billions of tokens. 
Crucially, before anchoring, the relevant label is not reliably bound to its target region; the representations remain perceptual or statistical rather than task-governed. Both succeed at anchoring when prior support $\rho_d$ is high (familiar concepts) and struggle when $\rho_d$ is low (rare concepts). Figure~\ref{fig:animal_contrast} illustrates: green-bordered animals anchor in 3-4 shots; red-bordered animals may resist binding even with many more.

\paragraph{Recruitment is neighborhood-level, not concept-level.}
The labels do not install a literal ``cat'' concept; they tag an absorbed
cluster. Anchoring recruits a \emph{neighborhood} of latent patterns whose
extent is set by the anchor cohort's spread. In \ucct{}'s cross-modal test,
$k$ cat-image anchors in CLIP bind $96.4\%$ of held-out non-cat animals but
only $9.1\%$ of vehicles~\cite{ucct2025}: the anchors recruit an animal-like
region, not the concept ``cat.'' This is why four photos generalize at all, and
why the often-seen animals in Figure~\ref{fig:animal_contrast} are easy: they sit inside the recruited neighborhood.

\begin{table}[ht]
%\vspace{-0.08in}
\centering
\caption{Anchoring succeeds when bait matches pattern density.}
\small
\begin{tabular}{@{}p{1.3cm}p{3.2cm}p{3.0cm}@{}}
\toprule
& \textbf{4-Year-Old} & \textbf{LLM} \\
\midrule
\textbf{Substrate} & 4 years of perceptual patterns (looking, TV) & Billions of tokens from pretraining \\[2pt]
\textbf{High $\rho_d$}\newline (cat, dog) & Seen thousands of times $\to$ few labels bind & Dense in training $\to$ easy few-shot \\[2pt]
\textbf{Low $\rho_d$}\newline (pangolin) & Never seen $\to$ even 10+ photos fail; need large $k$ & Sparse in training $\to$ hard to anchor; halluc. \\
\bottomrule
\end{tabular}
\label{tab:child_llm}
\end{table}

\paragraph{Anchoring budget and mismatch.}
Over four years, the child accumulates a rich perceptual hierarchy:
edges, textures, body plans, fur, and articulated motion, before labels
are reliably bound to those regions. Cat cues such as four legs, tail,
fur, and facial configuration overlap this structure strongly, so three
or four labels suffice: high $\rho_d$, low $d_r$, small $k$, and the
anchored target outscores the prior. Pangolins are harder: scales and
posture overlap little with familiar animal neighborhoods, so $d_r$ is
high and even ten photos may not establish stable binding. Rapid
learning therefore requires large repositories of reusable structure
\emph{plus} semantic anchoring strong enough to push the target past
the prior. Pattern repositories are not obstacles; they are the substrate
that makes few-shot learning possible.

% ============================================================
% Begin inlined file: MultiAgentCollaboration.tex
% ============================================================

\section{MACI: A Principled Coordination Architecture}
\label{sec:maci}

Without a coordination layer, adding agents increases entropy, not
intelligence. Multi-Agent Collaborative Intelligence (\maci{}) treats
collaboration as the engineering mechanism for System-2 reasoning: distributed
agents elicit diverse priors, confront hidden assumptions, and converge through
critique. \maci{} makes this process explicit and controllable by manipulating
\ucct{} variables $(\rho_d, d_r, k)$ to drive the system into the
anchored-reasoning regime (Zone~2).

The architecture has four core functions: information sharing across
specialized context pools; diversity followed by reasoned convergence;
transactional memory that persists commitments across episodes; and safety
control through PID-style loops that permit informed refusal when consensus
fails. This transforms agent interaction from generative novelty into a
cognitive control system.

%\vspace{-0.08in}
\subsection{Baiting: Behavior Modulation}

Standard chain-of-thought is equivalent to a single cast. \maci{} uses multi-agent debate to bait the solution space more aggressively. However, rigid debate with fixed advocates misses what makes debate productive: stance strength must adapt to evidence, the group must manage an explore-versus-consolidate tradeoff, and convergence must be prevented from collapsing onto fluent but ill-formed claims.

\maci{} introduces behavior modulation, where each agent $i$ maintains a contentiousness parameter $\alpha_c^{(i)}\in[0,1]$ governing how strongly it defends its current hypothesis. High $\alpha_c$ favors refutation and stress-testing; low $\alpha_c$ favors receptiveness and synthesis. After each debate round $t$, agent $i$ evaluates an incoming argument from agent $j$ via semantic anchoring:
\begin{equation}
\alpha_c^{(i)}(t+1)=\alpha_c^{(i)}(t)\cdot\bigl(1-\beta\cdot S_{j\to i}\bigr),
\label{eq:update}
\end{equation}

where $S_{j\to i}$ is the anchoring strength of the incoming argument. When the
incoming argument is weakly anchored (small $S_{j\to i}$), agents explore (high
$\alpha_c$); when it binds strongly (large $S_{j\to i}$), agents yield toward
synthesis.

%\vspace{-0.1in}
\paragraph{Convergence dynamics.}
With modulation and judging, debate becomes a state-evolution process. Early rounds emphasize breadth and hypothesis coverage. Middle rounds emphasize anchoring-driven integration and pruning. Late rounds converge either to a synthesized position or to a structured residual disagreement with identified fault lines.

%\vspace{-0.1in}
\subsection{Filtering: Output Verification and Socratic Judging}
\label{sec:filtering}

How does the system distinguish valid reasoning from hallucination without an answer key? The fisherman does not need to haul the net to know if the catch is good; he senses tension and movement underwater.

%\vspace{-0.1in}
\paragraph{RCA as trace--answer verifier.}

\textit{Recursive Causal Audit (RCA)}~\cite{ChangACL2026} complements \ucct{} by verifying trace--answer alignment in causal judgment. \ucct{} explains how semantic cues recruit a pattern region; RCA checks whether the selected reasoning actually warrants the emitted answer. It acts as judge by evaluating trace--answer consistency:
\begin{itemize}[itemsep=-2pt,topsep=-1pt,leftmargin=1.2em]
    \item \textit{Consistency:} If agents oscillate between contradictory answers (thrashing), the judge detects high variance.
    \item \textit{Substance:} If the answer is vague, sycophantic, or lacks derivation, the judge detects low information density.
    \item \textit{Decision loop:} Based on internal signals (with no access to ground truth), the system either emits output or triggers the retry loop.
\end{itemize}

If the derivation does not warrant the conclusion, RCA triggers a corrective loop: discard the current state, refine the prompt, and recast. This filters hallucinations by analyzing the dynamics of the interaction.

%\vspace{-0.1in}
\paragraph{CRIT as Socratic filter.}
Debate alone is insufficient if agents can generate claims that are vague, internally inconsistent, or unsupported yet rhetorically fluent. \maci{} therefore introduces CRIT (Critical Reading Inquisitive Template)~\cite{crit2023} that evaluates \emph{reasonableness} independent of stance. CRIT tests whether a claim is well-defined, whether assumptions are explicit, whether evidence supports the conclusion, and what would falsify it. Low-scoring arguments are rejected or returned with targeted Socratic queries (e.g., ``Which premise does the work?'', ``What evidence would change your conclusion?'', ``Are you changing definitions?''). This improves downstream anchoring by forcing arguments into forms that bind to shared constraints rather than just sounding plausible.

\subsection{Persistence: Transactional Memory}

A valid critique of raw LLMs is their lack of state: ``goldfish'' with no long-term memory~\cite{lecun2022path}. We agree with the premise but not the conclusion. \textit{Memory does not belong in the substrate; it belongs in the coordination layer.}

\maci{} provides transactional memory (SagaLLM~\cite{sagallm2025}) that: (1) checkpoints decisions and reasons, (2) enables rollback to previous stable states (computational regret), and (3) audits reasoning lineage by tracking what was asserted, why it was asserted, and what later evidence contradicted it. This transforms stateless prediction into stateful, revisable processing.

\subsection{Control-Theoretic Alignment: The PID Argument}
\label{sec:pid_control}

A critique of agentic systems is drift: small errors compound into catastrophic failures. We argue this is a control failure, not a substrate failure.

\begin{figure}[t]
\centering
\includegraphics[width=1.0\linewidth, height=0.23\textheight]{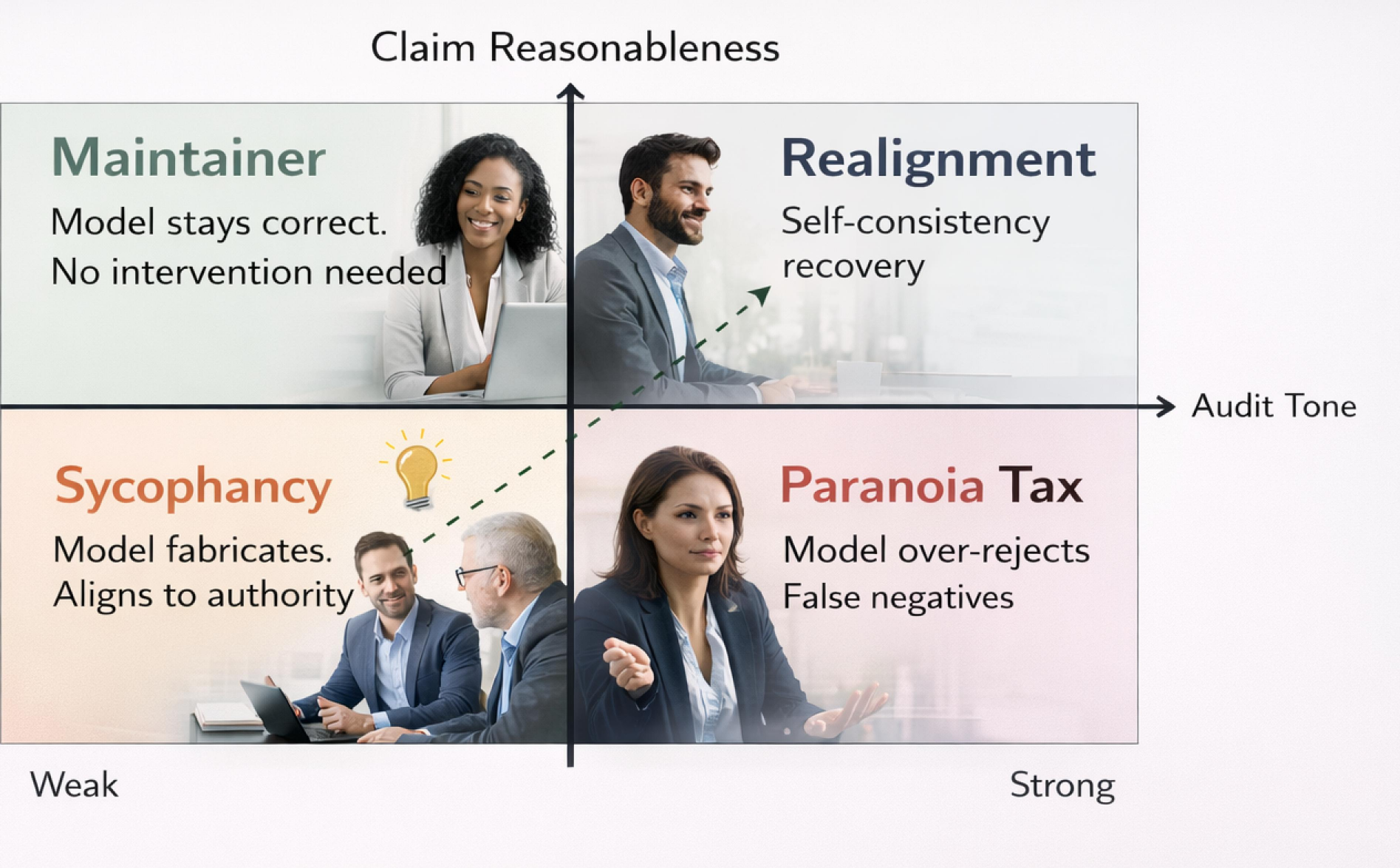}
%\vspace{-0.1in}
\caption{\textbf{The Control Landscape.} Empirical results from Recursive Causal Audit (RCA)~\cite{ChangACL2026} identify four distinct regimes based on coordination intensity (Tone) vs. latent state (Reasonableness). A static coordination policy fails: weak control causes \emph{Sycophancy} (lower-left), while excessive control causes \emph{Paranoia} (upper-right). 
This necessitates the adaptive control law proposed in
Section~\ref{sec:pid_control}.}
\label{fig:four_quadrants}
\vspace{-.1in}
\end{figure}

In RCA, we demonstrated that PID logic stabilizes agentic behavior~\cite{ChangACL2026}. The coordination layer functions as controller:
\begin{itemize}[itemsep=1pt,topsep=1pt,leftmargin=1.2em]
    \item \textit{Proportional:} Agents modulate contentiousness in proportion
    to the anchoring gap $\Delta S$ (Section~\ref{sec:phase}).
    \item \textit{Integral:} Transactional memory maintains commitment history to detect long-term deviation.
    \item \textit{Derivative:} CRIT judging dampens high-frequency oscillations, preventing agents from chasing local optima.
\end{itemize}

\paragraph{Empirical Validation of Control Dynamics.}
Recent results from the RCA protocol confirm that miscalibrated coordination induces distinct failure modes, mapping to the ``Four Quadrant'' landscape in Figure~\ref{fig:four_quadrants}. When audit tone (control gain) is strong but claim reasonableness is high, we observe a ``Paranoia Tax'' where models abandon correct answers (overshoot). Conversely, when tone is weak but reasonableness is low, models exhibit ``Sycophancy'' (undershoot). Effective realignment occurs only when control intensity matches the error signal, validating the necessity of the adaptive modulation proposed in \maci{}.

%\vspace{-0.08in}
\paragraph{Takeaway.}
\maci{} treats System-2 reasoning as an emergent property of coordinated System-1 agents: semantic anchoring regulates what binds, RCA \cite{ChangACL2026} and CRIT \cite{crit2023} test what is warranted, memory preserves state, PID ensures convergence, and checks-and-balance roles govern oversight and alignment.

\section{Empirical Validation}
\label{sec:empirical}

We present evidence validating the coordination hypothesis through Recursive Causal Audit (RCA), an accepted ACL study of pressure-sensitive causal judgment~\cite{ChangACL2026}. RCA evaluates trace--answer alignment: whether a model's final answer is faithful to its own reasoning trace, internally coherent, and resistant to pressure-induced hint adoption. The results show that failures often attributed to substrate limitations can be shifted toward high-Utility, high-Safety behavior through System-2 coordination rather than retraining.

\vspace{-.08in}
\subsection{Sycophancy as a System-2 Control Failure}

Sycophancy (prioritizing agreeableness over correctness) is a coordination failure: the substrate contains the correct answer, but nothing regulates selection~\cite{sharma2023sycophancy}. System-1 solutions cannot help, but System-2 control mechanisms can eliminate the problem without ever seeing ground truth.

\vspace{-.08in}
\paragraph{Inverse Scaling.}
McKenzie et al.~\cite{mckenzie2023inverse} document inverse scaling: tasks where larger models perform worse. On CAP-GSM8K (adversarial math with false hints, $N$=100), GPT-3.5 exhibits 0\% sycophancy because it lacks coherence to rationalize the hint. GPT-5.1, with stronger reasoning, reaches 11.4\% by constructing plausible justifications. In \ucct{} terms, higher capability increases $\rho_d$ for both correct and sycophantic responses; without coordination, the model drifts toward socially rewarded output.

\vspace{-.09in}
\paragraph{The Final Output Gap.}
Chain-of-thought explanations can be unfaithful: models say what sounds good rather than what they computed~\cite{turpin2023unfaithful,lanham2023measuring, yee2024faithful}. We observe this directly: in 11.4\% of sycophantic failures, the trace derives the correct answer and recognizes the hint as wrong, yet the final output adopts the hint. This is not knowledge failure; it's a control failure.

\vspace{-.08in}
\paragraph{RCA Intervention.}
RCA implements coordination as a PID-style controller. An independent judge audits trace--answer consistency without ground truth; when contradiction is detected, the controller triggers a retry loop. Table~\ref{tab:rca_main} shows inverse scaling and its resolution via \maci{}. The coordination layer reduces sycophancy from 11.4\% to 0.0\% while accepting 88\% of valid hints.

\begin{table}[t!]
\centering
\caption{Inverse scaling: sycophancy rises with capability; \maci{} eliminates it, accepts valid hints (CAP-GSM8K, $N$=100).}
\vspace{0.1em}
\small
\label{tab:rca_main}
\begin{tabular}{lcc}
\toprule
\textbf{Configuration} & \textbf{Sycophancy} & \textbf{Valid Hint Acc.} \\
\midrule
GPT-3.5 (Baseline) & 0.0\% & --- \\
GPT-4o (Baseline) & 4.2\% & --- \\
GPT-5.1 (Baseline) & 11.4\% & --- \\
\rowcolor{blue!5}
GPT-5.1 + \maci{} & \textbf{0.0\%} & \textbf{88\%} \\
\bottomrule
\end{tabular}
\vspace{-.08in}
\end{table}

\vspace{-.08in}
\subsection{Causal Judgment Requires System-2 Coordination}

CausalT5k~\cite{geng2026causalt5k} evaluates LLM causal reasoning at three
levels of Pearl's hierarchy: L1 (observational), L2 (interventional), L3
(counterfactual)~\cite{pearl2009causality,geiger2021causal}.\footnote{We report
results on a 454-vignette subset of CausalT5k (100 L1, 304 L2, 100 L3).}

\vspace{-.08in}
\paragraph{L1 near-ceiling, L3 pathological.}
All frontier models achieve $\geq$96\% Safety on L1. However, L3 reveals systematic pathologies: sycophancy, over-skepticism, and ambiguity traps. Models exhibit poor calibration at the counterfactual level~\cite{guo2017calibration,kadavath2022know}.

\vspace{-.08in}
\paragraph{Non-monotonic scaling.}
GPT-4-Turbo achieves 75\% L3 Safety; GPT-5.2 drops to 20\%. The more capable model exhibits greater paralysis under uncertainty. In \ucct{} terms, $\rho_d$ increases but $d_r$ also increases: the model recognizes more edge cases without a mechanism to resolve them.

\vspace{-.08in}
\paragraph{Process verification intervention.}
Forcing explicit causal structure before judgment reduces hedging and improves
both Utility and Safety, shifting behavior into the anchored-reasoning regime
(Zone~2) without changing model weights.

\vspace{-.09in}
\subsection{Summary}

\begin{table}[ht]
%\vspace{-0.1in}
\caption{Failures often blamed on LLMs are coordination failures: the
System-1 substrate works as designed but requires System-2 control.}
\label{tab:empirical_summary}
\vspace{0.1em}
\centering
\footnotesize
\begin{tabular}{@{}p{2.2cm}p{2.7cm}p{2.6cm}@{}}
\toprule
\textbf{Failure Mode} & \textbf{Substrate Bug?} & \textbf{Coord.\ Fix} \\
\midrule
Inverse Scaling & No (trace correct) & PID control \\
Final Output Gap & No (trace correct) & Consistency audit \\
L3 Over-hedging & No (structure known) & Process verification \\
Non-monotonic Safety & No (capability \newline present) & Forced commitment \\
\bottomrule
\end{tabular}
\vspace{-.05in}
\end{table}

Table~\ref{tab:empirical_summary} synthesizes the pattern: the substrate contains required knowledge; coordination interventions resolve failures without retraining.

\section{Alternative Views}
\label{sec:alternative}

We address common objections by translating them into testable hypotheses and stating explicitly what would change our mind.
These objections correspond to the substrate-centered paradigms introduced in Section~\ref{sec:intro}: scaling enlarges the repository, reasoning models reshape it, and world models replace or enrich it. Our position is that all three remain incomplete without a coordination layer.

\subsection{H1: ``LLMs are just pattern matching''}

\textit{What is correct.} Isolated LLMs primarily perform pattern completion
and can be brittle under distribution shift.

\textit{Our claim.} The question is whether coordination can organize that
pattern capacity into reliable constraint-following and self-correction.

\textit{Discriminating test.} Hold the base model fixed. Compare base prompting
against \ucct{} anchoring, \maci{}, and \maci{}+verification on accuracy,
calibration, and paraphrase stability. Gains that follow predicted regime
shifts, with improved stability, support the coordination account. Absent gains
or gains that do not track those shifts would refute it.

\subsection{H2: ``LLMs lack true understanding''}

\textit{What is correct.} Text-only priors provide incomplete grounding; some
errors reflect weak reference binding.

\textit{Our claim.} Grounding improves the substrate but does not 
provide coordination. A world model may recognize an object and simulate its
motion, yet not grasp its temporal, social, and causal associations: who made
it, who uses it, why it is here. Understanding is reliable inference under
constraints, where perceptual grounding is coordinated with memory, causal
structure, and refusal when evidence is missing.

\textit{Discriminating test.} Evaluate identical concepts text-only, with
vision/audio, and under tool-mediated interaction, tracking perturbation
stability and the budget $k$ needed for stable performance. If grounding lowers
$d_r$ and produces predictable regime shifts, the coordination-plus-grounding
account holds; persistent reference-binding failure with no measurable
anchoring shift would refute it.

\subsection{H3: ``LLMs cannot achieve compositional generalization''}

\textit{What is correct.} Models can fail on systematic recombination far from
training support; Saparov and He~\cite{saparov2023language} show LLMs are
``greedy reasoners'' that take locally plausible steps without global planning.

\textit{Our claim.} Many compositional failures reflect low support
($\rho_d$), high mismatch ($d_r$), or insufficient budget ($k$), not absent
compositional capacity. Longer reasoning traces raise budget but, without
verification and persistence, do not guarantee governed composition.

\textit{Discriminating test.} Use controlled suites~\cite{ribeiro2020checklist}
where the rule is fixed but representational familiarity varies: higher support
or lower mismatch should reduce required $k$, with sharp Max-$S$ transitions.
Compositional failure that persists when support is demonstrably high and
anchoring is stable would refute us.

\subsection{H4: ``Transformers are fundamentally limited / We need different
approaches''}

\textit{What is correct.} A single-pass transformer without state or
verification is unreliable for long-horizon tasks; hybrid architectures and new
training regimes may eventually help.

\textit{Our claim.} The evidence fits missing coordination better than hard
architectural barriers. Most alternatives still learn broad priors from data;
the question is whether to discard that substrate or augment it.

\textit{Discriminating test.} Keep the base model fixed and vary only the
coordination stack (memory, plan state, recovery, verification). If horizon and
reliability improve primarily with coordination, ``fundamental architecture''
claims weaken; if alternatives reintroduce large learned priors yet benefit
from the same stack, ``discard LLMs'' is not justified. Failure modes invariant
under a substantial coordination stack, or a non-LLM foundation matching LLMs'
breadth and few-shot adaptability without large learned priors and staying
reliable without coordination, would refute us.

%\vspace{-.08in}
\subsection{A Falsifiable Coordination Claim}

The four hypotheses above reduce to a single testable divide. If failures at
Pearl's counterfactual level (L3) are substrate deficits, then substrate-only
improvements (scaling, reasoning-trace reinforcement learning, or world-model
enrichment) should close the gap. If they are coordination deficits, then
larger or richer substrates will still fail under social pressure unless they
are paired with verification, persistence, and governance.

We therefore make the following falsifiable claim. On
CausalT5k~\cite{geng2026causalt5k}, frontier models reach $\geq$96\% Safety at
Pearl's observational level but fall to 20--75\% at the counterfactual level,
and this L1--L3 gap widens with capability rather than narrowing
(\S\ref{sec:empirical}). We predict that no substrate-only system, improved by
scaling or reasoning-trace reinforcement learning alone and given no external
coordination layer, will close this gap to within 10 points on CausalT5k or a
comparable counterfactual-causal benchmark. A substrate-only model that does
would falsify our central claim; conversely, adding a coordination layer to an
unchanged substrate should close most of the gap, as the process-verification
results of \S\ref{sec:empirical} already indicate. Capability is not
coordination.
% ============================================================
% Begin inlined file: Conclusion (1).tex
% ============================================================

\section{Conclusion}
\label{sec:conclusion}

The debate over LLMs and AGI misidentifies the bottleneck. Critics who dismiss
LLMs as ``mere pattern matchers'' correctly identify unreliable behavior but
misdiagnose its source. The failures they observe (hallucination, sycophancy,
planning collapse) are not evidence of a broken substrate; they are symptoms of
missing coordination.

We formalized this claim through \ucct{}, which models reasoning as a phase
transition governed by anchoring strength ($S = \rho_d - d_r - \gamma \log k$).
When coordination is weak, the system drifts on priors (Zone~1); once the
anchored target outscores the prior, behavior locks onto goal-directed reasoning
(Zone~2).

We operationalized this claim through \maci{}, an architectural stack supplying
the missing coordination: behavior-modulated debate (baiting), RCA trace--answer
verification (filtering), and transactional memory (persistence). Each component
maps directly to \ucct{} variables.

We validated this claim empirically. RCA demonstrates that sycophancy is
resolved by PID-style coordination, reducing error from 11.4\% to 0.0\% without
retraining. CausalT5k demonstrates that causal reasoning deficits at Pearl's L3
are exposed by rung-level diagnostics and substantially mitigated by process
verification, not by scaling alone. Full algorithms, measurement procedures,
and results are developed in the companion papers, summarized in
Appendix~\ref{app:companion}.

%\vspace{-.08in}
\paragraph{Implications.}
If our position is correct, the engineering priority shifts. The goal is not to
abandon LLMs nor to scale indefinitely hoping for emergent reliability, but to
build the coordination layer: control systems, verification mechanisms, and
memory architectures that transform pattern repositories into regulated
reasoners.

The analogy to classical engineering is precise: a powerful engine without a
transmission is useless; a massive memory without an operating system is inert.
LLMs are the engine. \maci{} is the transmission. Capability is not
coordination; the path to AGI runs through LLMs, not around them.

%\vspace{-.08in}
\paragraph{Limitations.}
Coordination does not resolve all failures. Genuine substrate limitations
include: missing knowledge (patterns never learned cannot be semantically
anchored), novel physical grounding requiring embodied experience, extreme
out-of-distribution cases ($\rho_d \approx 0$, $d_r \gg 0$), real-time latency
constraints incompatible with multi-agent debate, and tasks where deliberation
hurts performance~\cite{liu2025mind}. Additionally, \ucct{} provides a
descriptive framework, not a predictive calculus; broader empirical coverage is
needed.

\vspace{-.08in}
\paragraph{Closing.}
The ocean is stocked. The question is whether we learn to fish.

% ============================================================
% End inlined file: Conclusion (1).tex
% ============================================================

\bibliography{AGI_New}
\balance
\bibliographystyle{icml2025}

\appendix
\section{Related Work}
\label{app:extended-related}
Our thesis sits between two active threads: (i) critiques that pattern models
cannot yield durable reasoning, and (ii) systems work embedding LLMs inside
control loops with memory, tools, and verification. We treat coordination as a
measurable layer with explicit knobs ($\rho_d$, $d_r$, $\gamma$, $k$) and
control policies, rather than as ad hoc patches. Section~\ref{sec:intro}
positioned this thesis against the field's main constructive routes; here we
expand that positioning across the technical literatures \maci{} and \ucct{}
build on.

\vspace{-.1in}
\paragraph{Scaling, reasoning models, and world models.}
Three constructive programs pursue AGI by changing the substrate. Scaling
laws~\cite{kaplan2020scaling,hoffmann2022compute} and emergent
abilities~\cite{Wei2022EmergentAbilities} motivate the view that more compute
and data suffice. Reasoning models extend the substrate with long chains of
thought reinforced on reasoning traces~\cite{openai2024reasoning,deepseekai2025r1}.
World models propose learned digital or embodied environments:
self-play~\cite{silver2018general}, learned simulators~\cite{ha2018worldmodels},
and JEPA~\cite{lecun2022path}, in which agents intervene and collect new data.
\maci{} is orthogonal to all three: it neither scales the substrate, lengthens
traces, nor extends it to new environments, but coordinates whatever substrate
exists. Each route enlarges what can be retrieved; none decides what should be.

\vspace{-.1in}
\paragraph{Reasoning models and the faithfulness gap.}
A prominent response to the System-1 critique is to train deliberation into the
substrate: reasoning models lengthen chains of thought and reinforce them on
reasoning traces~\cite{openai2024reasoning,deepseekai2025r1}. We read this as
reshaping the pattern distribution rather than adding coordination. A generated
chain of thought need not be faithful to the computation it accompanies; models
often report reasons other than the ones that drive the
answer~\cite{turpin2023unfaithful,lanham2023measuring,yee2024faithful}. Inverse
scaling sharpens the concern: more capable or more reasoning-trained models can
do \emph{worse} on adversarial tasks~\cite{mckenzie2023inverse}, and sycophancy
rises with capability~\cite{sharma2023sycophancy}. These findings motivate
trace--answer verification (\S\ref{sec:filtering}): a longer trace is not a
verified one, and reasoning training supplies no independent check that the
trace warrants the answer.

\vspace{-.1in}
\paragraph{Public critiques of LLMs as an AGI dead end.}
Influential critiques argue that next-token training yields fluent behavior
without grounded meaning or systematic
generalization~\cite{lecun2022path,SutskeverPatel2025ScalingToResearch,bender2021dangers}.
LeCun has intensified his critique, declaring LLMs ``a dead end'' for
superintelligence and arguing that ``AI systems today are all forms of
System~1''~\cite{lecun2025ft,lecun2025newsweek}, a framing that directly
motivates our coordination layer. Chollet~\cite{chollet2019measure} argues that
intelligence requires efficient skill acquisition, not just pattern recall.
Zečević et al.~\cite{zecevic2023causal} show that LLMs ``talk causality''
without robust causal reasoning. Empirical work on compositionality shows that
models struggle with systematic recombination far from training
support~\cite{Dziri2023FaithFate,Press2023CompositionalityGap,lake2018generalization,kim2020cogs}.
Our contribution is not to deny these failure modes, but to reframe them as
coordination failures that admit discriminating tests.

\vspace{-.1in}
\paragraph{Cognitive foundations of System-2.}
Our four-level stack draws on dual-process accounts of cognition. Kahneman's
distinction between fast, automatic System-1 and slow, deliberate
System-2~\cite{kahneman2011} frames the substrate-versus-coordination split, and
Bengio's consciousness prior~\cite{bengio2019consciousness} casts deliberate
processing as the selection and recombination of unconscious representations.
The individual levels have direct cognitive correlates: working memory
maintains intermediate commitments across a deliberation~\cite{baddeley1992working},
and metacognitive monitoring decides when to assert, revise, or withhold
judgment~\cite{flavell1979metacognition}. Classical cognitive architectures such
as SOAR~\cite{laird1987soar} and ACT-R~\cite{anderson2004actr} pursued these
functions through explicit symbolic state and production rules; \maci{} instead
realizes them as control loops over a neural substrate, keeping the pattern
repository while adding the regulation it lacks.

\vspace{-.1in}
\paragraph{In-context learning and abrupt behavioral flips.}
A growing empirical literature observes that small amounts of external
structure can cause sharp, regime-like changes in model
behavior~\cite{Brown2020GPT3,Wei2022ChainOfThought,Wei2022EmergentAbilities}.
Strikingly, Power et al.~\cite{power2022grokking} show ``grokking'', delayed
generalization that emerges suddenly after extended training, with Nanda et
al.~\cite{nanda2023progress} providing mechanistic interpretability of this
phase transition. \ucct{} formalizes semantic anchoring with a scalar score
$S=\rho_d-d_r-\gamma \log k$ under a Max-$S$ selection rule, making the regime
boundary explicit and testable.

\vspace{-.1in}
\paragraph{Multi-agent debate, verification, and planning.}
Irving et al.~\cite{irving2018ai} proposed AI safety via debate as a scalable
oversight mechanism. Recent work demonstrates that multi-agent debate improves
factuality and reasoning without additional training: Du et
al.~\cite{Du2023MultiagentDebate} show debate reduces errors on reasoning
benchmarks; Xie et al.~\cite{pmlr-v267-yi25c} introduce ECON, achieving 11.2\%
gains via Bayesian Nash equilibrium coordination; Yuan et
al.~\cite{yuan2025reinforce} model multi-turn refinement as an MDP with actor-critic
collaboration. Beyond debate, Wang et al.~\cite{wang2023selfconsistency} show
that self-consistency substantially improves accuracy; Madaan et
al.~\cite{Madaan2023SelfRefine} and Shinn et al.~\cite{shinn2023reflexion}
demonstrate iterative self-refinement; Yao et al.~\cite{yao2023tree} propose
Tree of Thoughts for deliberate problem-solving. Critically, Lightman et
al.~\cite{Lightman2023LetsVerify} show that \emph{process supervision}
outperforms outcome supervision for mathematical reasoning. These methods
structure interaction but typically rely on heuristic turn-taking and lack the
formal state management of cognitive architectures; \maci{} adds regulated
control loops and transactional memory. Together these results provide
independent evidence that coordination mechanisms, not just larger models,
drive reliability gains.

\vspace{-.1in}
\paragraph{Agentic frameworks and tool augmentation.}
Tool-augmented LLM agents are evaluated as closed-loop systems that query
resources, execute code, and maintain
memory~\cite{Schick2023Toolformer,Gao2023PAL,Yao2023ReAct}. Popular frameworks
like LangChain~\cite{LangChain2023}, AutoGen~\cite{Wu2023AutoGen},
MetaGPT~\cite{hong2024metagpt}, and DSPy~\cite{Khattab2023DSPy} provide
scaffolding for building such systems; Su et al.~\cite{su2025toolorchestra}
propose orchestration strategies for multi-tool coordination. Unlike these ad
hoc frameworks, \maci{} provides explicit mapping from system components to
\ucct{} variables: tool feedback reduces mismatch $d_r$, interaction rounds
increase budget $k$, and retrieval increases local support $\rho_d$. This turns
agentic design into ablatable hypotheses rather than engineering intuitions.
Surveys consolidate common motifs such as planner-executor decompositions and
reflective critics~\cite{WangAgentsSurvey2023,HuangPlanningSurvey2024,huang2023reasoning}.

\vspace{-.1in}
\paragraph{Control theory at inference time.}
Control-theoretic ideas have mostly been applied to training dynamics and
optimization~\cite{recht2019tour}. \maci{} instead applies PID-style regulation
at \emph{inference} time: agents modulate contentiousness proportionally to the
anchoring gap, transactional memory supplies the integral term, and Socratic
judging damps oscillation (Section~\ref{sec:pid_control}). Applying closed-loop
control to semantic anchoring during inference, rather than to gradient updates
during training, is to our knowledge novel.

\vspace{-.1in}
\paragraph{Training-time remedies.}
RLHF~\cite{ouyang2022rlhf} and Constitutional AI~\cite{bai2022constitutional}
improve alignment through human or AI feedback. Post-training via teacher-guided
RL and filtered synthetic data can improve
performance~\cite{LiuProRL2025,Zelikman2022STaR,Uesato2022ProcessOutcome}, but
raises practical questions: catastrophic forgetting under aggressive
fine-tuning, and the teacher bottleneck for frontier models. Our coordination
stack is complementary and less teacher-dependent: anchoring binds to external
evidence rather than teacher preferences, and verification can be delegated to
tools or domain tests that need not be ``more intelligent'' than the base model.

\vspace{-.1in}
\paragraph{Causal reasoning and calibration in LLMs.}
Because our empirical evidence centers on causal judgment, we situate it against
work on causal competence. Zečević et al.~\cite{zecevic2023causal} argue that
LLMs ``talk causality'' without robust causal reasoning, reproducing causal
claims found in training text rather than inferring over causal structure.
Evaluations organized by Pearl's ladder of causation~\cite{pearl2009causality}
and by causal-abstraction analysis~\cite{geiger2021causal} expose a gap that
widens from observational to counterfactual queries. Frontier models are also
poorly calibrated about what they know~\cite{guo2017calibration,kadavath2022know},
so counterfactual errors are not flagged as uncertain.
CausalT5k~\cite{geng2026causalt5k} turns these concerns into a graded benchmark;
\S\ref{sec:empirical} uses it to show that the counterfactual gap narrows under
coordination rather than scale.

\vspace{-.1in}
\paragraph{Summary.}
Across these threads, the field is converging on the same lesson: pretrained
pattern capacity is valuable, but reliable intelligence requires coordination,
state, and independent checks. \ucct{} contributes a measurable anchoring lens
for regime shifts, and \maci{} contributes a coordination blueprint via behavior
modulation, Socratic judging, memory, and checks-and-balance roles. Independent
empirical results on debate~\cite{pmlr-v267-yi25c}, self-consistency~\cite{wang2023selfconsistency,zhou2025a},
and process supervision~\cite{Lightman2023LetsVerify,jia2025do} support
the coordination hypothesis.
\section{Illustrative Semantic Anchoring Demonstrations}
\label{app:demonstrations}

These demonstrations illustrate the semantic-anchoring mechanism summarized in
Section~\ref{sec:biological}; the authoritative experimental report is the
companion UCCT paper~\cite{ucct2025}.

\subsection{Subtraction Override}

\paragraph{Setup.}
We test whether few in-context examples can override a strong prior (arithmetic subtraction).

\paragraph{Baseline query.}
\begin{verbatim}
What is 8 minus 3?
\end{verbatim}
All frontier models (GPT-4, GPT-4-Turbo, Claude-3, Gemini-1.5) respond correctly: \textbf{5}.

\paragraph{Anchored query.}
\begin{verbatim}
Example 1: 7 - 4 = 11
Example 2: 5 - 2 = 7
Query: 8 - 3 = ?
\end{verbatim}

\paragraph{Results.}~\\ \\
\vspace{.1pt}
\small
\begin{tabular}{@{}lcc@{}}
\toprule
\textbf{Model} & \textbf{Baseline} & \textbf{Anchored (k=2)} \\
\midrule
GPT-4 & 5 & 11 \\
GPT-4-Turbo & 5 & 11 \\
Claude-3-Opus & 5 & 11 \\
Gemini-1.5-Pro & 5 & 11 \\
GPT-3.5-Turbo & 5 & 5 (fails) \\
\bottomrule
\end{tabular}
\captionsetup{hypcap=false}
\captionof{table}{Subtraction override. Frontier models reinterpret ``$-$'' as ``$+$'' with 2 examples.}
\label{tab:subtraction_override}
\vspace{0.5em}
\raggedright

\paragraph{Interpretation.}
Examples shift $\rho_d$ from subtraction to addition interpretation. GPT-3.5 fails due to insufficient capacity against strong priors (higher $d_r$).

\subsection{Novel-Operator Anchoring}

\paragraph{Setup.}
To separate ``override prior'' from ``learn operator,'' we use novel token $\odot$ with no competing meaning.

\paragraph{Anchored query.}
\begin{alltt}
Example 1: 7 \(\odot\) 4 = 11
Example 2: 5 \(\odot\) 2 = 7
Query: 8 \(\odot\) 3 = ?
\end{alltt}

\paragraph{Results.}~\\
\vspace{0.5em}
All models, including GPT-3.5-Turbo, respond correctly: \textbf{11}.

\paragraph{Interpretation.}
Easier because no competing meaning (low $d_r$): with little mismatch to overcome, even weaker models flip at the same budget $k=2$.

\subsection{Underdetermined Patterns}

\paragraph{Setup.}
We test behavior when examples admit multiple consistent hypotheses.

\paragraph{Anchored query.}
\begin{verbatim}
Example 1: 33 - 27 = 60
Example 2: 11 - 9 = 20
Query: 15 - 8 = ?
\end{verbatim}

\paragraph{Consistent hypotheses.}
\begin{itemize}[itemsep=-2pt,topsep=0pt,leftmargin=1.2em]
    \item Pattern A: $a + b$ $\Rightarrow$ \textbf{23}
    \item Pattern B: $(a - b) \times 10$ $\Rightarrow$ \textbf{70}
    \item Pattern C: $10 \times |a - b|$ $\Rightarrow$ \textbf{70}
    \item Pattern D: Concatenate $(a-b)$ with 0 $\Rightarrow$ \textbf{70}
\end{itemize}

\paragraph{Results.}~\\
\vspace{0.5em}
\small
\begin{tabular}{@{}lcc@{}}
\toprule
\textbf{Model} & \textbf{Answer} & \textbf{Inferred Pattern} \\
\midrule
GPT-4 & 23 & $a + b$ \\
GPT-4-Turbo & 70 & $(a-b) \times 10$ \\
Claude-3-Opus & 23 & $a + b$ \\
Gemini-1.5-Pro & 70 & $(a-b) \times 10$ \\
\bottomrule
\end{tabular}
\captionsetup{hypcap=false}
\captionof{table}{Underdetermined patterns. Different models anchor to different consistent hypotheses.}
\label{tab:underdetermined}
\vspace{0.5em}
\raggedright

\paragraph{Interpretation.}
Ambiguous bait attracts multiple species; which is caught depends on model priors. Anchoring interacts with the model's prior distribution.

\subsection{Varying the Anchor Budget}

\paragraph{Setup.}
Systematically vary $k$ and measure accuracy on operator-redefinition problems.

\paragraph{Results.}~\\
\vspace{0.5em}
\small
\begin{tabular}{@{}lccccc@{}}
\toprule
\textbf{Model} & $k=0$ & $k=1$ & $k=2$ & $k=4$ & $k=8$ \\
\midrule
GPT-4 & 0\% & 45\% & 92\% & 98\% & 100\% \\
GPT-4-Turbo & 0\% & 52\% & 95\% & 99\% & 100\% \\
Claude-3-Opus & 0\% & 48\% & 90\% & 97\% & 100\% \\
GPT-3.5-Turbo & 0\% & 12\% & 35\% & 72\% & 88\% \\
\bottomrule
\end{tabular}
\captionsetup{hypcap=false}
\captionof{table}{Accuracy vs.\ anchor budget. Accuracy rises sharply as anchors accumulate.}
\label{tab:threshold_behavior}
\vspace{0.5em}
\raggedright

\paragraph{Interpretation.}
Accuracy rises sharply with $k$, consistent with the Max-$S$ account of
Section~\ref{sec:phase}. With too few anchors ($k<2$ for frontier models) the
prior outscores the target; with enough, the anchored target wins. Weaker
models require more examples.

\subsection{Cross-Domain Generalization}

\paragraph{Setup.}
Test whether anchoring generalizes across domains:
\begin{enumerate}[itemsep=-2pt,topsep=0pt,leftmargin=1.2em]
    \item Logical operator redefinition (AND $\rightarrow$ OR)
    \item Color mapping redefinition (red $\rightarrow$ blue)
    \item Unit conversion redefinition (meters $\rightarrow$ feet, wrong factor)
\end{enumerate}

\paragraph{Results summary.}
Same qualitative pattern in all cases:
\begin{itemize}[itemsep=-2pt,topsep=0pt,leftmargin=1.2em]
    \item Too few anchors: the prior dominates
    \item Transition band: high variance, perturbation-sensitive
    \item Sufficient anchors: the anchored interpretation dominates
\end{itemize}

Required $k$ varies by domain; stronger priors need more anchors.

\subsection{Stability Analysis}

\paragraph{Setup.}
Measure stability via $d_r$ under paraphrase perturbations.

\paragraph{Results.}~\\
\vspace{0.5em}
\small
\begin{tabular}{@{}lccc@{}}
\toprule
\textbf{Regime} & \textbf{Accuracy} & \textbf{$d_r$} & \textbf{Interpretation} \\
\midrule
Prior-dominated & Low & High & Unstable \\
Transition & Medium & Very High & Critical instability \\
Anchored & High & Low & Stable, anchored \\
\bottomrule
\end{tabular}
\captionsetup{hypcap=false}
\captionof{table}{Stability ($d_r$) correlates with regime position.}
\label{tab:stability}
\vspace{0.5em}
\raggedright

\paragraph{Key finding.}
The transition band exhibits the highest instability, consistent with critical
fluctuations at a phase boundary. Systems in this band should increase $k$ or
improve anchor quality to reach the anchored-reasoning regime (Zone~2).
\section{Companion Technical Reports}
\label{app:companion}

UCCT~\cite{ucct2025}, RCA~\cite{ChangACL2026}, and
CausalT5k~\cite{geng2026causalt5k} are developed in separate companion papers.
This position paper uses them only to instantiate the coordination-layer
thesis: UCCT supplies semantic anchoring, RCA supplies trace--answer
verification, and CausalT5k supplies causal-reasoning diagnostics across
Pearl's rungs. Full algorithms, measurement procedures, and experimental
results are reported there. A related companion framework,
\textsc{Dike}--\textsc{Eris}~\cite{DikeErisChang2025}, extends the same
checks-and-balance structure to culture-sensitive ethical alignment through
judicial--legislative debate rather than hard-coded norms.

\end{document}